\documentclass[12pt]{article}
\usepackage{graphicx} 
\usepackage{mathptmx}
\usepackage{etoolbox}
\AfterEndEnvironment{figure}{\vskip-12pt}

\usepackage[a4paper,  
    left=30mm,
    top=25mm,
    right=25mm,
    bottom=25mm,
    ]{geometry}

\begin{document}

\centering

\hfill \break
\hfill \break

\textbf{COMPARING MEASURES OF LINGUISTIC DIVERSITY ACROSS SOCIAL MEDIA LANGUAGE DATA AND CENSUS DATA AT SUBNATIONAL GEOGRAPHIC AREAS}\\
\bigskip
\bigskip

\raggedright
\textbf{Sidney Gig-Jan Wong, Jonathan Dunn, Benjamin Adams}\\
\bigskip

\raggedright
University of Canterbury, New Zealand\\
sidney.wong@pg.canterbury.ac.nz, jonathan.dunn@canterbury.ac.nz,
benjamin.adams@canterbury.ac.nz
\bigskip
\bigskip

\raggedright
Abstract
\bigskip

\raggedright
This paper describes a preliminary study on the comparative linguistic ecology of online spaces (i.e., social media language data) and real-world spaces in Aotearoa New Zealand (i.e., subnational administrative areas). We compare measures of linguistic diversity between these different spaces and discuss how social media users align with real-world populations. The results from the current study suggests that there is potential to use online social media language data to observe spatial and temporal changes in linguistic diversity at subnational geographic areas; however, further work is required to understand how well social media represents real-world behaviour.
\bigskip
\bigskip

\raggedright
INTRODUCTION
\bigskip

\raggedright
Linguists are often challenged by limited language data when attempting to observe and identify language variation and change in a population. Samples from a small number of individuals do not necessarily represent the larger community in the aggregate. Computational linguistic methods bypass the limitations of traditional methods in sociolinguistics and dialectology which rely on limited language samples and small multivariate datasets (Nguyen et al. 2016).
\bigskip

\raggedright
Social media platforms such as Twitter are a rich source of language data needed to test linguistic theories on written language. There are 4.63 million internet users in Aotearoa New Zealand in January 2022 which is equivalent to an internet penetration rate 94.9\% (Kemp, 2022). However, Dunn et al. (2020) argues that online language data is inherently skewed. This means we need to account for biases when making claims about linguistic phenomena about an underlying population. This is because the linguistic ecologies of the two spaces will differ.
\bigskip

\raggedright
If we consider the user base of different social media platforms, Facebook reported 3.05 million Aotearoa New Zealand-based users compared with Twitter which has a smaller user base of 552,000 users as of 2022 (Kemp, 2022). This means Twitter only represents\\ 13.6\% of eligible users aged 13 years and over with access to the internet in Aotearoa New Zealand (Kemp, 2022). This means the linguistic ecologies between online and real-world may not be directly comparable.
\bigskip

\raggedright
Dunn et al. (2020) noted that we can correct for non-local bias in samples of online social media language data by using measures of linguistic diversity. They identified changes in linguistic diversity over the course of the COVID-19 pandemic for each country as nationwide lockdowns limited international and domestic movement (Dunn et al., 2020). Not only were they able to identify and correct for non-local populations at a national level; the results found that the linguistic ecology of online spaces changed as a result of real-world events (Dunn et al., 2020).
\bigskip

\raggedright
This paper addresses two questions: firstly, what are the similarities between language ecologies in real-world and online/social media at subnational geographies; and secondly, what insights can we obtain from analysing social media language data at subnational geographies.
\bigskip
\bigskip

\raggedright
METHODOLOGY
\bigskip

\raggedright
Data Sources
\bigskip

\raggedright
The data for this study comes from two sources: 1) the georeferenced Twitter sub-corpus from the web-based Corpus of Global Language Use (CGLU; Dunn, 2020) and 2) subnational demographic information from the New Zealand Census of Population and Dwellings (Census; Stats NZ 2019).
\bigskip

\raggedright
A major benefit of using social media language data from Twitter is the volume of language data produced by users. This allows us to observe temporal and spatial variation across measures of linguistic diversity. 
\bigskip

\raggedright
Only tweets originating from Aotearoa New Zealand were included in our current analysis. Each tweet is georeferenced and coded to a population centre across 98 locations within a 50-kilometre radius from the given coordinates (Dunn, 2020). The earliest collected tweet was dated 2017-07 and the most recent tweets were dated 2022-04.
\bigskip

\raggedright
We compared the Twitter language data with the ``languages spoken'' question from the Census. Stats NZ does not collect written language ability in the census. The question asks: ``In which language(s) could you have a conversation about a lot of everyday things?'' which is a multiple response variable. 
\bigskip

\raggedright
The Census is a national count of all individuals in Aotearoa New Zealand; therefore, we consider it a suitable data source to calculate ground truth subnational measures of linguistic diversity. We did not include the categories `New Zealand Sign Language', `Other', and `None (e.g. too young to talk)' in our analysis. 
\bigskip
\bigskip

\raggedright
Data Processing
\bigskip

\raggedright
\textit{Language Identification}
\bigskip

\raggedright
We used language identification classifiers to automatically code the primary language of a Tweet.
\pagebreak

\raggedright
Two classifiers are included as part of our analysis for comparability as there may be classification errors (i.e., as a result of codeswitching). The first language classifier `LID1' uses the `idNet' package Dunn (2020) while the second language classifier `LID2' is adapted to identify Austronesian languages (e.g., te reo Māori) (Dunn and Nijhof; 2022).
\bigskip

\raggedright
\textit{Geographic Concordance}
\bigskip

\raggedright
We validated the online language data to ensure that the information only relates to Aotearoa New Zealand. There were data points from 100 locations across Aotearoa New Zealand. Each location corresponds with a geohash. 
\bigskip

\raggedright
We then linked the Tweets to the 16 regional council areas. The rationale for using regional council areas as opposed to customised areas is so that we are able to combine our aggregated Twitter data to published Census data. 
\bigskip

\raggedright
\textit{Linguistic Diversity Measures}
\bigskip

\raggedright
We used a concentration ratio (CR) as a proxy for linguistic diversity (Hirschman 1945). A CR is used to determine the market structure and competitiveness of the market and provides a range between 0\% to 100\%. The CR is calculated as follows (1):
\bigskip

\raggedright
(1) CR$n$ = C1 + C2 + ... + C$n$
\bigskip

\raggedright
Where:
\bigskip

\raggedright
C$_{n}$ defines the share of the nth largest languages as a percentage of a population\\
$n$ defines the number of languages included in the CR calculation
\bigskip

\raggedright
We use a 10-firm CR (CR$_{10}$). Common CR measures include 4-firm (CR$_{4}$) and 8-firm (CR$_{8}$).  A CR$_{10}$ between 0\% to 40\% suggests low concentration (i.e., perfect competition). A CR$_{10}$ between 40\% to 70\% suggests medium concentration (i.e., an oligopoly). A CR$_{10}$ between 70\% to 100\% suggests high concentration (i.e., a monopoly).
\bigskip
\bigskip

\raggedright
RESULTS
\bigskip

\raggedright
The initial measure of the national CR$_{10}$ figures for the 2006 Census was 0.76, 2013 Census was 0.81, and the 2018 Census was 0.79. The two-year delay between 2006 and 2013 censuses was a result of the 2010 and 2011 Canterbury Earthquakes.
\bigskip

\raggedright
The CR$_{10}$ measures from the Twitter corpus from the 2018 year found that the LID1 classifier was 0.79 and the LID2 classifier was 0.72. The initial results suggest that Aotearoa New Zealand is typically linguistically homogeneous. The LID1 classifier was more consistent with the 2018 Census which we could use as a baseline measure.
\bigskip

\pagebreak

\centering
\textbf{Table 1}: CR$_{10}$s for 2018 by regional council areas
\bigskip

{
\centering
    \begin{tabular}{|l|l|c|c|c|c|}\hline
        Region Code & Region Name & 2018 Census & LID1 & LID2 & Number of Tweets\\\hline
         01 & Northland & 0.76 & 0.56 & 0.52 & 32,976\\\hline
         02 & Auckland & 0.60 & 0.79 & 0.73 & 195,182\\\hline
         03 & Waikato & 0.76 & 0.81 & 0.75 & 260,518\\\hline
         04 & Bay of Plenty & 0.75 & 0.80 & 0.73 & 40,734\\\hline
         05 & Gisborne & 0.70 & 0.88 & 0.48 & 8,162\\\hline
         06 & Hawke's Bay & 0.78 & 0.95 & 0.86 & 23,777\\\hline
         07 & Taranaki & 0.85 & 0.49 & 0.44 & 26,341\\\hline
         08 & Manawatū-Wanganui & 0.80 & 0.71 & 0.66 & 69,809\\\hline
         09 & Wellington & 0.72 & 0.90 & 0.83 & 151,881\\\hline
         12 & West Coast & 0.89 & 0.66 & 0.60 & 14,393\\\hline
         13 & Canterbury & 0.80 & 0.79 & 0.72 & 141,939\\\hline
         14 & Otago & 0.82 & 0.93 & 0.87 & 40,398\\\hline
         15 & Southland & 0.87 & 0.89 & 0.81 & 21,375\\\hline
         16 & Tasman & 0.86 & 0.58 & 0.54 & 43,061\\\hline
         18 & Marlborough & 0.85 & 0.83 & 0.75 & 31,377\\\hline
     \end{tabular}
}
\bigskip

\raggedright
In Table 1, we provide the CR$_{10}$ measures for Census and Twitter data from the 2018 year by regional council areas. The Census CR$_{10}$ measures are based on the total responses of individuals (i.e., one individual may indicate one or more languages on the Census form) while the CR$_{10}$ measures calculated from the language classification models (LID1 and LID2) are based on Tweets.
\bigskip

\raggedright
These subnational measures show great variability unlike the national measures. Every regional council area differed in measures of linguistic diversity between the 2018 Census. Only Canterbury, Southland, and Marlborough had similar levels of linguistic diversity between the 2018 Census and LID1 classifier measures. The LID2 classifier is lower than both the 2018 Census and LID1 classifier except for Auckland, Hawke's Bay, Wellington, and Otago. There is a consistent negative relationship between measures of linguistic diversity between Census and Twitter; the correlation coefficient for LID1 was -0.29 and for LID2 was -0.10. The correlation coefficient between LID1 and LID2 was 0.80 which suggests a positive relationship.
\bigskip

\centering
\textbf{Table 2}: Top 10 language varieties for the 2018 year
\bigskip

{
\centering
    \begin{tabular}{|c|c|c|c|}\hline
        Rank & 2018 Census & LID1 & LID2\\\hline
        1 & English & English & English\\\hline
        2 & Māori & Portuguese & Portuguese\\\hline
        3 & Samoan & Japanese & Thai\\\hline
        4 & Northern Chinese & Tagalog & Japanese\\\hline
        5 & Hindi & Spanish & Spanish\\\hline
        6 & French & Indonesian & Tagalog\\\hline
        7 & Yue & Arabic & Malaysian\\\hline
        8 & Sinitic nfd & French & French\\\hline
        9 & Tagalog & Korean & Arabic\\\hline
        10 & German & Thai & Korean\\\hline
     \end{tabular}
}

\raggedright
The results from Table 1 raises the question: how similar are the 2018 Census and the Twitter corpus? Table 2 compares the top 10 language varieties from the 2018 Census and the Twitter corpus at a national level. We expected English to rank highly, but the distribution of languages between the two data sources differs significantly. We also note none of the Chinese languages (e.g., Northern Chinese, Yue, or Sinitic not further defined (nfd)) are in the top 10 list. Twitter is officially banned in the People’s Republic of China which may account for the smaller user base of Chinese languages on Twitter.
\bigskip

\begin{figure}[h!]
    \includegraphics[width=15.5cm, height=6.71cm]{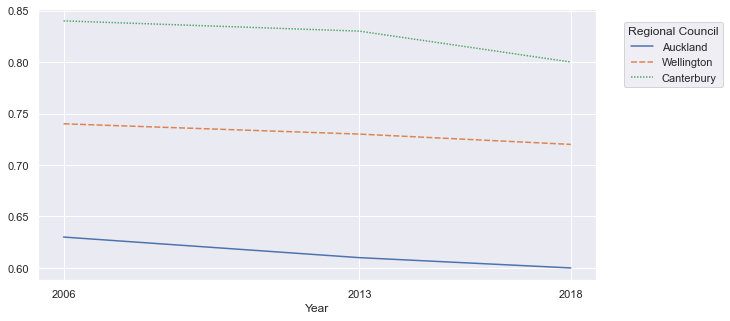}
\end{figure}
\centering
\textbf{Figure 1}: CR$_{10}$ measures from Census for select regional council areas
\bigskip

\raggedright
For the purposes of this paper, we concentrated our analyses on Auckland, Wellington, and Canterbury regional council areas. Figure 1 is a line graph of the CR$_{10}$ measures from Census over three census cycles. We can see from the Census data that the CR$_{10}$ is decreasing which suggests linguistic diversity is increasing between census cycles. Auckland is the most linguistically diverse followed by Wellington and Canterbury Regions.

\begin{figure}[h!]
    \includegraphics[width=15.5cm, height=6.66cm]{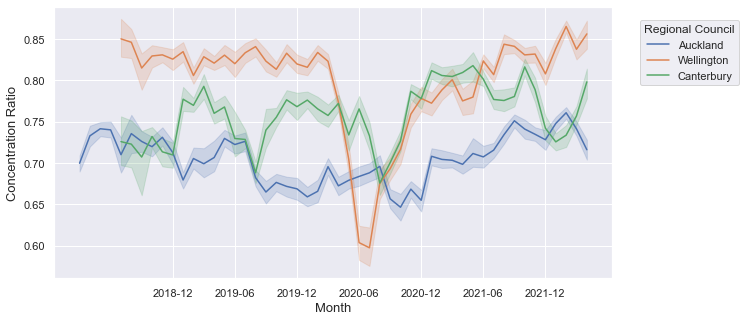}
\end{figure} 
\centering
\textbf{Figure 2}: CR$_{10}$ measures from Twitter by month for select regional council areas
\bigskip

\raggedright
We expected this pattern to persist in the Twitter corpus. Figure 2 is a line graph of the CR$_{10}$ measures from Twitter by month with the upper and lower 95\% percentile confidence intervals included for each area. We have restricted our analysis to the LID2 classifier as it provides higher accuracy when classifying languages from the Austronesian language family. We can observe high levels of variation of linguistic diversity over time for all three regional council areas; however, the trends are more opaque than the 2018 Census.\\ We still observe Auckland as having the highest level of linguistic diversity. We can also observe a massive increase in linguistic diversity for the Wellington region around June 2020. This was unexpected.
\bigskip

\begin{figure}[h!]
    \includegraphics[width=15.5cm, height=6.66cm]{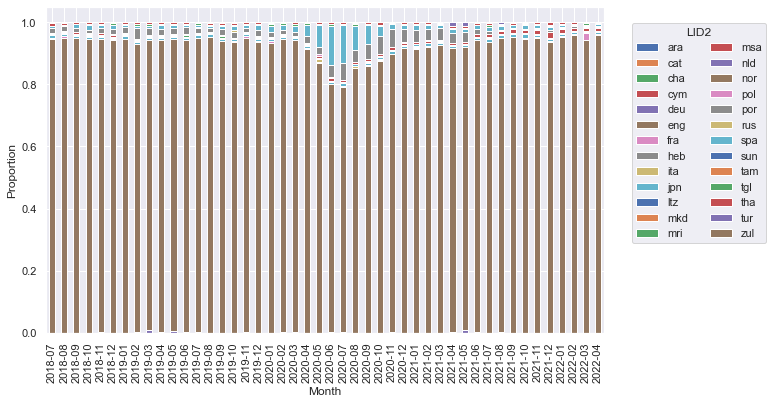}
\end{figure}
\centering
\textbf{Figure 3}: Proportion of languages from Twitter by month for Wellington
\bigskip

\raggedright
Figure 3 shows a stacked bar graph of the proportion of languages from Twitter using the LID2 classifier. We only included the top ten language varieties for each month. We can observe an increase of Spanish and Portuguese in the period with increased linguistic diversity between April 2020 and September 2020. This period coincided with the beginning of the national lockdown due to the on-going Covid-19 pandemic. Covid-19 was known as `corona' and there is a possibility that the language identification classifiers considered tweets with this word as Spanish. However, removing the two strings did not have any impact on the CR$_{10}$ or proportion of languages for Wellington.
\bigskip
\bigskip

\raggedright
DISCUSSION AND CONCLUSION
\bigskip

\raggedright
The results from the current study suggests that we can use online social media language data to observe spatial and temporal changes in linguistic diversity for subnational geographic areas; however, we should be cautious in how we interpret trends we observe online.
\bigskip

\raggedright
The high CR$_{10}$ from the census could be the result of English being included in the measure. English is the de facto official language of Aotearoa New Zealand due to colonisation. Based on the latest figures from the 2018 Census, approximately 95.7\% of the usually resident population of Aotearoa New Zealand can speak English which suggests many respondents to the Census who can speak one or more languages will be bilingual.
\bigskip

\raggedright
The nature of written language differs significantly from spoken language as there is not always a one-to-one relationship between the two modes of language (in some cases there is a zero-to-one, one-to-many, or many-to-many relationships between modes). This means language varieties represented in the 2018 Census data may not appear in the Twitter\\ corpus and vice versa. If we compare the list of languages in each data source, there were up to 403 distinct language categories from the LID1 and LID2 classifiers, while there\\ were only 196 categories in the Language classification from the Census recognised by Stats NZ.
\bigskip

\raggedright
The geographic spread of locations across Aotearoa New Zealand raises questions about data quality from a spatial perspective. With reference to Table 1, tweets for the Nelson region were missing. It is possible tweets originating from the Nelson Region have been absorbed into neighbouring Tasman and Marlborough regions due to the data collection methodology as they are within the 50-kilometre radius of each other.
\bigskip

\raggedright
The information available from Twitter, and in extension online and social media, provides a rich source of language data for us to observe real-time societal and attitudinal changes. With reference to Figure 3, we can observe changes in the socio-political landscape on social media because of current events. For example, we can see a small increase of Arabic language tweets in March 2019 and May 2019 following the terrorist attack on Christchurch masjidain on 15 March 2019. 
\bigskip

\raggedright
Before we can compare language ecologies between the real-world and social media, we still need to determine the relationship between the two spaces. This limits the conclusions we can draw from our current analysis without knowing the true population of interest in online spaces. Despite these challenges, the current short paper provides promising results as Census data only provides a snapshot of a location at a specific time point, whereas online and social media data provides contemporaneous information in a small sample of the population. 
\bigskip
\bigskip

\raggedright
REFERENCES
\bigskip

\raggedright
Dunn J, 2020. Mapping languages: the Corpus of Global Language Use. \textit{Lang Resources \& Evaluation} 54, 999–1018.
\bigskip

\raggedright
Dunn J and B Adams, 2019. Mapping Languages and Demographics with Georeferenced Corpora, \textit{Proceedings of Geocomputation 2019}.
\bigskip

\raggedright
Dunn J, T Coupe, and B Adams, 2020. Measuring Linguistic Diversity During COVID-19, \textit{Proceedings of The Fourth Workshop on the Fourth Workshop on Natural Language Processing and Computational Social Science}.
\bigskip

\raggedright
Dunn J, and W Nijhof, 2022. Language Identification for Austronesian Languages, \textit{13th International Conference on Language Resources and Evaluation}. 
\bigskip

\raggedright
Albert O Hirschman. 1945. \textit{National power and the structure of foreign trade}. Univ of California Press. 
\bigskip

\raggedright
Kemp S, 2022, \textit{Digital 2022: New Zealand}, Web site: https://datareportal.com/reports/digital-2022-new-zealand (accessed May 1, 2022).
\bigskip

\raggedright
Nguyen D, AS Doğruöz, CP Rosé, and F de Jong, 2016. Computational Sociolinguistics: A Survey. \textit{Computational Linguistics} 42, 537–593. 
\pagebreak

\raggedright
Stats NZ, 2019, \textit{Languages spoken (total responses) and birthplace (broad geographic areas) by age group and sex, for the census usually resident population count, 2006, 2013, and 2018 Censuses (RC, TA, DHB)}, Web site: https://nzdotstat.stats.govt.nz/wbos/Index.aspx?DataSetCode=TABLECODE8286 (accessed May 1, 2022). 
\bigskip

\raggedright
Keywords
\bigskip

\raggedright
linguistic diversity, social media, census
\bigskip

\end{document}